# Text Embedded Swin-UMamba for DeepLesion Segmentation

Ruida Cheng[a], Tejas Sudharshan Mathai[c], Pritam Mukherjee[c], Benjamin Hou[b], Qingqing Zhu[b], Zhiyong Lu[b], Matthew McAuliffe[a], Ronald M. Summers[c]

[a] Scientific Application Services, Center of Information Technology, NIH
[b] Division of Intramural Research (DIR), National Library of Medicine (NLM), NIH
[c] Imaging Biomarkers and Computer-Aided Diagnosis Laboratory, Radiology and Imaging Sciences, Clinical Center, NIH

## ABSTRACT

Segmentation of lesions on CT enables automatic measurement for clinical assessment of chronic diseases (e.g., lymphoma). Integrating large language models (LLMs) into the lesion segmentation workflow has the potential to combine imaging features with descriptions of lesion characteristics from the radiology reports. In this study, we investigate the feasibility of integrating text into the Swin-UMamba architecture for the task of lesion segmentation. The publicly available ULS23 DeepLesion dataset was used along with short-form descriptions of the findings from the reports. On the test dataset, our method achieved a high Dice score of 82.64 ± 17.36% and a low Hausdorff distance of 6.34 ± 10.48 pixels was obtained for lesion segmentation. The proposed Text-Swin-U/Mamba model outperformed prior approaches: 37.79% improvement over the LLM-driven LanGuideMedSeg model ($p < 0.001$), and surpassed the purely image-based XLSTM-UNet and nnUNet models by 2.58% and 1.01%, respectively. The dataset and code can be accessed at https://github.com/ruida/LLM-Swin-UMamba

## 1. INTRODUCTION

Lesion assessment on CT is essential for early diagnosis, treatment planning, and longitudinal monitoring of chronic diseases, such as lung and liver cancer. [1,2] In current clinical practice, lesion size and characteristics (e.g., enhancement, attenuation attributes, irregularity) play a predominant role in the determination of malignancy of suspicious findings[3,4,5]. For example, the management of liver lesions follows the sizing guidelines of the American College of Gastroenterology[1], while lung nodules are to be managed using an entirely different set of sizing guidelines from the Fleischner Society[2]. Accordingly, radiologists measure the long- and short axis diameters (LAD and SAD) of lesions, which serves as surrogate marker for malignancy. Moreover, it is cumbersome for radiologists to manually size the dimensions of several lesions during the busy clinical day, especially when faced with metastasis. Other confounding elements are the acquisition of CT studies on myriad scanners from different manufacturers and the diverse appearances and shape of lesions in these studies. Despite these variations, radiologists must dictate the size and characteristics of suspicious lesions into a report.

While several prior works for lesion segmentation have been proposed, few have investigated the utility of lesion descriptions present in the reports. In natural image domain, many prior works have incorporated Large Language Models (LLMs) for image segmentation or object detection tasks.[6,7,8] However, porting their work to the medical image domain poses a significant challenge due to suboptimal segmentations as organs or tumors often have lower contrast, ambiguous boundaries, and complex anatomical relationships. Our approach does not rely on existing mainstream foundation models, such as ChatGPT[9], because they are not open-sourced and cannot directly be trained or fine-tuned for task-specific medical image segmentation. An early study, Liu et al.[10] explored a Contrastive Language Image Pre-training(CLIP)-driven universal model to tackle CLIP embedding for organ labels. Li et al.[11] reported a text-augmented Transformer model for segmentation, L-ViT, while Zhong et al.[12] used GuideDecoder and text prompts to enhance the segmentation results. All three architectures, in general, encoded text into the segmentation backbone, either guiding or improving the prediction of segmentation masks. Importantly, all three approaches missed the inclusion of a text decoder component due to the complexity of the relational memory-based architecture.[13] Previous works by Yan et al.[14,15] combined lesion features and short-form report text descriptions to learn relationships among related lesions, attributes, and surrounding anatomy.

In this study, we investigate the feasibility of adapting text-based embedding directly for lesion segmentation. We present the Text-Swin-UMamba model that implements multiple language-embedding mechanisms (Text Tower[19] and LLaVA-style[23] encoder) to embed short-form report text descriptions into the decoder of Swin-UMamba[14] for lesion segmentation. Our method was compared against prior approaches (LanGuideMedSeg[13], xLSTM-UNet[20], and nnUNet[21]) and demonstrated improved lesion segmentation performance through fusion of text and imaging features.

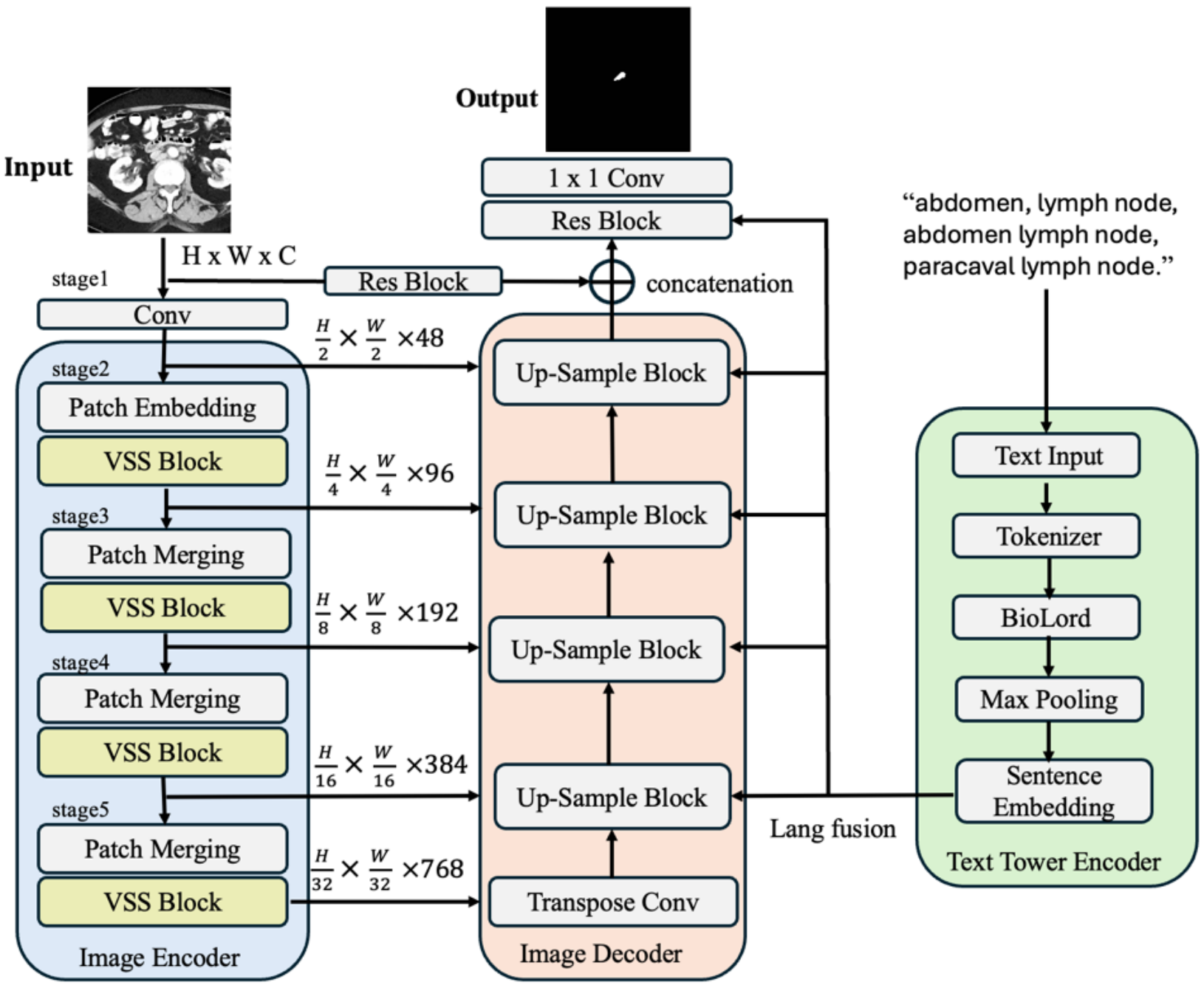

Figure 1. The design of Text-Swin-UMamba where the Text Tower encodes short-form descriptions of clinical findings on 2D slices in the DeepLesion dataset and embeds them into the Swin-UMamba segmentation backbone.

## 2. METHODS

The overview of the proposed Text-Swin-UMamba architecture is shown in Figure 1. The model is composed of three major building blocks: the Swin-UMamba image encoder, the Swin-UMamba image decoder, and the text tower encoder. We follow the Swin-UMamba implementation, a derivative of the fully automated and self-configuring nnU-Net framework, as the segmentation backbone of the proposed model. For the text embedding, we utilize the Text Tower[19] encoder to convert short-form text from the radiology reports into textual feature representations. The Text Tower building block aims to take a text report as input, tokenize it, pass it through a pre-trained "BioLord" model, and then produce a pooled (summarized) embedding for each text input. The Tokenizer layer serves as a custom text preprocessing utility. The BioLord pretrained model can produce meaningful representations for clinical sentences. The max pooling layer obtains a fixed-size embedding from the variable-length token embedding produced by the BioLord tokenizer. Max pooling selects, for each hidden dimension, the single token with the highest activation among all valid contextualized token embeddings. Max pooling keeps the strongest feature value across all tokens. The final output of the Text Tower is a fixed-size vector (e.g., 768 dimensions) that represents the semantic meaning of the entire input text report.

The language features are then integrated into the Swin-UMamba decoding path at multiple scales, particularly at five stages of the decoder, corresponding to feature maps (Figure. 1). The language feature is passed through a dedicated linear projection layer (Lang Fusion) to match the channel dimensions of the respective image features. The multimodal fusion mechanism between the Text Tower encoder and the Swin-UMamba decoder can guide the Swin-UMamba with text-based semantic information. A few other text encoding and fusion mechanisms are also implemented, such as LLaVA Text Tower, Bert-based encoding and fusion, and tail encoding (Please refer to Github repository for implementation detail).

## 3. EXPERIMENTS

**Dataset.** The ULS23 DeepLesion dataset[17] was utilized as validated labels were available in this dataset. The dataset tailored from the original DeepLesion dataset, which contained 4427 unique patients across 10,594 studies, and 32,120 CT image slices (age: 57 ±18 years, 56% males, 44% females). The short form text report description corresponding to a finding on a specific 2D slice in the original DeepLesion dataset (and annotated in the ULS23 dataset) was extracted. Each sentence provides detailed information, including its location (e.g., left lung, retroperitoneum), lesion type (e.g., nodule, mass, lymph node), and relevant attributes (e.g., hypoattenuation, enhancement, irregularity). The rationale for the choice of this dataset was due to the limited availability of datasets with annotated lesion and corresponding report text. For example, Qiu et al.[16] did not release their lesion benchmark dataset, SegLesion. Each 2D CT image slice is 512x512 pixels in size. A significant number of lesion masks appear in a small or tiny region of the image slice, posing substantial challenges for direct lesion segmentation. The 2D CT image slice, its corresponding mask label from the ULS23 DeepLesion dataset, and the short sentence description constituted our dataset, which was divided into training (n = 15040), validation (n = 3760), and testing (n = 1807) data subsets, respectively. The dataset was divided at the patient level with no overlap across the splits.

**Comparisons.** The proposed LLM-Swin-UMamba method was compared against prior approaches including LanGuideMedSeg [12], xLSTM-UNet [20], and 2D nnU-Net [21].

**Implementation Details.** Five-fold cross-validation is conducted in training and validation stage. All models are trained on a single NVIDIA V100x GPU with 32 GB of memory. The loss function integrates the Dice coefficient loss with the cross-entropy loss. AdamW optimizer with initial learning rate of 5e-3, scheduled via cosine anneal. The training epochs are set for 1500. The complete training phase for each model takes approximately 3 to 5 days to complete.

## 4. RESULTS

Table 1 describes the results of the various models on the test dataset, while Figure 2 shows the distribution of Dice scores. Figure 3 details qualitative segmentation results overlaid on the CT. The proposed Text-Swin-UMamba model improved the Dice score by 37.79 % (82.64 % vs. 44.85 %, $p < 0.001$) over the LanGuideMedSeg model. Against other models without text embedding, it outperformed the xLSTM-UNet model by 2.58% (82.64 % vs. 80.06 %, $p < 0.001$) and achieved a slightly higher mean Dice score of 1.01% ($p > 0.05$) compared to the 2D nnU-Net model. Notably, the model obtained a lower HD error than the 2D nnU-Net (6.34 vs. 6.86 pixels).

Table 1. Testing phase comparison of segmentation models performance with the DeepLesion dataset.

| | Mean Dice (%) | Jaccard (%) | Hausdorff95 (px) | Sensitivity (%) | Specificity (%) | Text Embed |
|---|---|---|---|---|---|---|
| LanGuideMedSeg [12] | 44.85 ± 24.18 | 32.20 ± 21.41 | 14.86 ± 15.29 | 47.10 ±.26.20 | 99.23 ± 0.77 | ✓ |
| xLSTM-UNET[20] | 80.06 ± 19.93 | 70.35 ± 22.05 | 7.52 ± 12.91 | 83.92 ± 16.08 | 99.76 ± 0.24 | × |
| nnUNet[21] | 81.63 ± 18.37 | 72.25 ± 21.27 | 6.86 ± 11.56 | 83.56 ± 16.44 | 99.77 ± 0.23 | × |
| Text-Swin-UMamba | 82.64 ± 17.36 | 73.49 ± 20.64 | 6.34 ± 10.48 | 84.60 ± 15.40 | 99.82 ± 0.18 | ✓ |

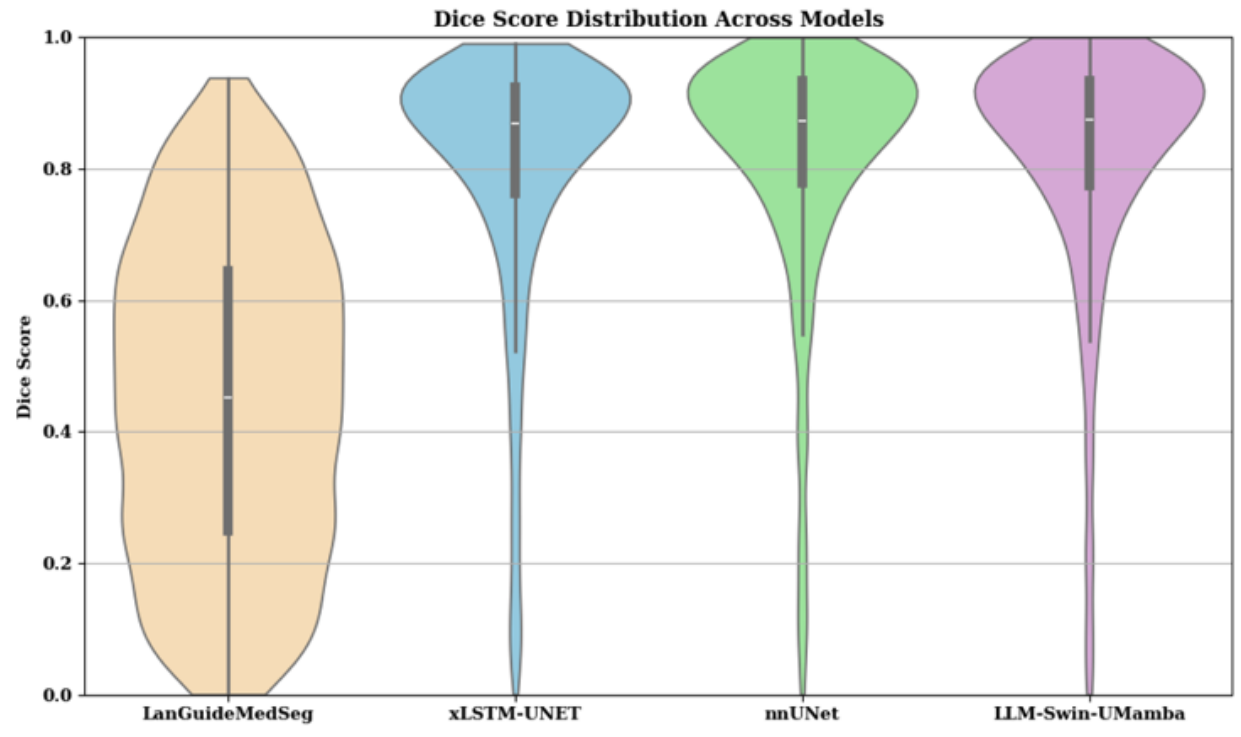

Figure 2. Distribution of Dice scores for each model on the ULS23 test dataset

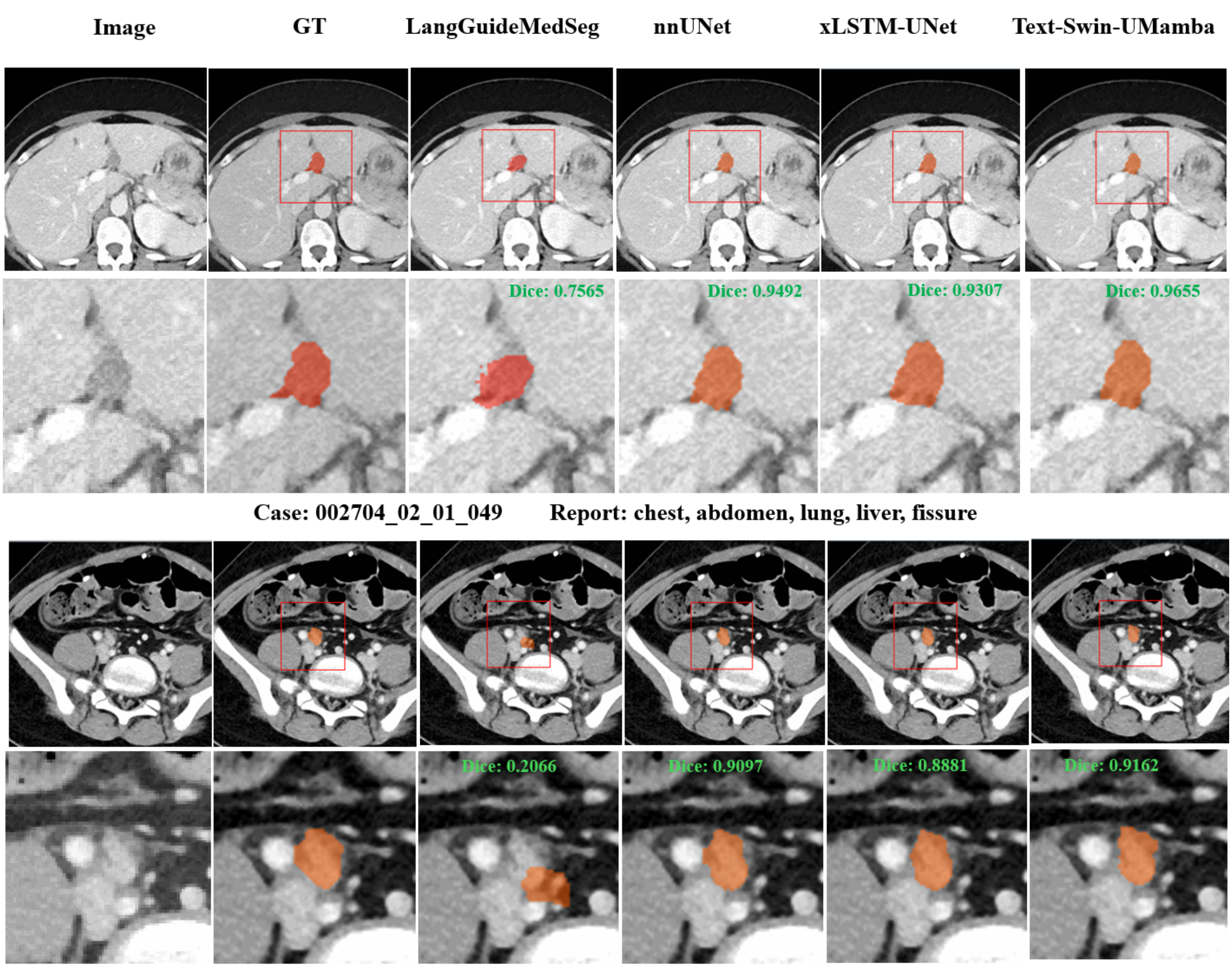

**Case: 001702_01_01_077** **Report: lymph node, pelvis, enhancing**

Figure 3. Qualitative results of segmentation of a liver lesion and a lymph node by the proposed Text-Swin-UMamba model compared against the other approaches.

## 5. ABLATION EXPRIMENTS

Motivated by Xing et al.[24], we compare the effect of different pooling mechanisms for token-level embeddings in the Text Tower encoder. We replaced the original mean-pooling layer with a max-pooling layer and a weighted-average pooling layer, evaluated the segmentation performance. Table 2 summarizes the comparison of the DeepLesion dataset.

Table 2. Comparison of segmentation model performance using different Text Tower pooling layers on the DeepLesion dataset.

| | Mean Dice (%) | Jaccard (%) | Hausdorff95 (px) | Sensitivity (%) | Specificity (%) | Text Embed |
|---|---|---|---|---|---|---|
| mean pooling | $82.34 \pm 17.66$ | $73.20 \pm 21.07$ | $6.45 \pm 10.99$ | $82.56 \pm 17.44$ | $99.83 \pm 0.17$ | ✓ |
| weight average pooling | $82.22 \pm 17.78$ | $73.04 \pm 21.16$ | $6.58 \pm 11.53$ | $82.12 \pm 17.88$ | $99.82 \pm 0.18$ | ✓ |
| max pooling | $82.64 \pm 17.36$ | $73.49 \pm 20.64$ | $6.34 \pm 10.48$ | $84.60 \pm 15.40$ | $99.82 \pm 0.18$ | ✓ |

The max pooling mechanism achieved slightly better performance among all the metrics. However, the differences among the three pooling layers were minor. The limited performance variation could be due to the nature of the short-form radiology report, which contains relatively few tokens and occupies a small fraction of memory after fusion; as a result, the different pooling methods may, in general, have less effect on the text embedding.

We further examined the impact of injecting the text-embedded features at all levels of the image decoder, including the bottleneck layer of the proposed Text-Swin-UMamba architecture. Table 3 presents the quantitative comparison between full-scale injection and injecting the text feature only at the decoder stage.

Table 3. Comparison of segmentation model performance using different text feature injections on the DeepLesion dataset.

| | Mean Dice (%) | Jaccard (%) | Hausdorff95 (px) | Sensitivity (%) | Specificity (%) | Text Embed |
|---|---|---|---|---|---|---|
| full injection | 82.75 $\pm$ 17.25 | 73.56 $\pm$ 20.42 | 6.46 $\pm$ 10.76 | 84.08 $\pm$ 15.92 | 99.79 $\pm$ 0.21 | ✓ |
| decoder only | 82.64 $\pm$ 17.36 | 73.49 $\pm$ 20.64 | 6.34 $\pm$ 10.48 | 84.60 $\pm$ 15.40 | 99.82 $\pm$ 0.18 | ✓ |

The full-scale text-feature injection conducts a slight 0.11% improvement in the mean Dice score over the decoder-only injection; however, the gain is marginal due to the short-form reports.

To further demonstrate the contribution of text information, we experiment to evaluate the difference between the original Swin-UMamba architecture (without text embedding) and the proposed Text-Swin-UMamba model. Table 4 exhibits the quantitative results.

Table 4. Comparison of segmentation model performance between Swin-UMamba and the proposed Text-Swin-UMamba on the DeepLesion dataset.

| | Mean Dice (%) | Jaccard (%) | Hausdorff95 (px) | Sensitivity (%) | Specificity (%) | Text Embed |
|---|---|---|---|---|---|---|
| Swin-UMamba | 81.53 $\pm$ 18.10 | 71.95 $\pm$ 20.91 | 6.72 $\pm$ 11.07 | 84.56 $\pm$ 15.44 | 99.77 $\pm$ 0.23 | × |
| Text-Swin-UMamba | 82.64 $\pm$ 17.36 | 73.49 $\pm$ 20.64 | 6.34 $\pm$ 10.48 | 84.60 $\pm$ 15.40 | 99.82 $\pm$ 0.18 | ✓ |

The Text-Swin-UMamba model achieves a 1.11% improvement in mean Dice score over the baseline Swin-UMamba architecture. Although the gain is moderate, the text-enhanced model outperforms the baseline on all evaluation metrics, indicating that incorporating short-form radiology report information could provide complementary cues that contribute to DeepLesion segmentation.

## 6. DISCUSSION

In this paper, we present the initial research phase of our study, with the ultimate goal of developing a unified framework that integrates LLM-based reasoning, lesion segmentation, bounding-box detection, and radiology report generation using the original DeepLesion dataset. In this initial phase, we evaluated the feasibility of embedding short-form diagnostic text into the Swin-UMamba architecture for lesion segmentation. A primary contribution of the present work is the demonstration that text-based semantic features can be incorporated into nnUNet-derived architectures using a simple Text Tower mechanism. At this stage, our approach does not yet include grounding models that explicitly localize anatomical regions referenced in the text.

Importantly, automatic text report generation has not yet been realized in nnUNet models in the current literature. The second phase of our research will therefore focus on encoding long-form radiology reports, developing text decoders, and enabling full report generation integrated with medical-imaging grounding models. The follow-up work will directly address how text and image information can operate together in the clinical workflow.

As noted in the introduction, only a limited number of studies have explored text embedding for medical imaging segmentation and classification. While prior works such as CLIP-driven embedding (Liu et al.10), L-ViT (Li et al.11), and GuideDecoder (Zhong et al.12) primarily embedded text into the segmentation backbone and did not include a text decoder because of the complexity associated with relational-memory architectures jointly trained with either the segmentation or detection backbone. Yan et al. 14,15 mapped each word in short-form reports to a specific number using one-hot encodings to predict reports, which is fundamentally different from LLM-based embedding. Therefore, text-guided approaches to enhance medical image segmentation or detection remain an active research area with substantial opportunities for advancement, not a closed problem.

In the second phase of our work, we plan to integrate long-form text embedding, lesion localization, and report generation into a single unified architecture. This unified system is intended to support and streamline clinical workflows for lesion diagnosis, with imaging findings (segmentation and object detection) and report generation complementing each other.

In our experiments, text embedding produced a modest improvement in segmentation performance—for example, a 1.01% increase in mean Dice score over the 2D nnUNet baseline. Although the performance gains over image-only models are modest, our analysis suggests that the text provides complementary semantic context that is not fully captured from 2D CT images alone. Short-form radiology descriptions typically include key semantic attributes—such as lesion type (e.g., nodule, lymph node), coarse anatomical location (e.g., left hepatic lobe, retroperitoneum), and qualitative characteristics (e.g., hypoattenuating, enhancing, irregular). In challenge cases of visual appearance in the surrounding anatomy region on a 2D CT image slice, these clinical cues could help reduce ambiguity. Thus, even brief text provides additional global information about the target lesion and its expected location, which helps stabilize segmentation predictions in challenging cases.

Several factors naturally limit the magnitude of improvement on the DeepLesion dataset. Short form reports contain limited detail. Each description is usually only one sentence with few tokens, providing coarse—not fine-grained—semantic information. Lesions in ULS23 are centered and cropped to 256×256 pixels from the surrounding region. This specific setting simplified the lesion-localization or search mechanism. The text-based cues provide only incremental benefit beyond the visual representation already embedded in the dataset design. Only 2D slices are used. Many radiologic descriptors (e.g., irregular margins, multilobulated, elongated) are inherently 3D concepts. When reduced to a single 2D slice, the synergy between text and imaging is constrained. Given these factors, we expect modest but consistent improvements—precisely what we observe.

Our qualitative analysis reveals that text integration in clinically meaningful scenarios could align closely with short-form reports in the DeepLesion dataset. The lesion type (i.e., lymph node, nodule, mass), the anatomical location (i.e., liver, lung, retroperitoneum), and the qualitative attributes (i.e., irregular, hypoattenuation, enhancement) can match the image appearance with the segmentation labels. Overall, the improvement patterns roughly resemble how radiologists use brief report text.

First, even small changes in Dice can significantly affect the clinical outcome of lesion segmentation, which is the primary tool for obtaining quantitative measurements such as long- and short-axis diameters, lesion volume, and changes over follow-up intervals. More accurate segmentation reduces differences in these measurements, which can affect cancer grading and follow-up. Thus, a slight increase in the Dice score can lead to better clinical measurements.

In practice, clinicians do not need precise contouring; they need segmentations that reliably capture lesion extent and shape sufficiently to support measurement and comparison across time. In general, different observers score manual lesion annotation with Dice coefficients usually above 0.80. Our model gets an average Dice of 82.55%, which is slightly better than typical human agreement for many CT lesions. Therefore, the segmentation quality already meets clinically acceptable performance standards.

Even though our study operates on 2D slices, inter- and intra-observer variation remains relevant. Different radiologists may outline different parts of the same lesion because the contrast is low, the boundary is ambiguous, or the radiologist's subjective interpretation of the lesion extent. This variance sets a realistic ceiling for the Dice metric that can be improved, suggesting that future enhancements might require more detailed text descriptions, 3D volumetric context, or advanced multimodal reasoning rather than just architectural changes.

Automated segmentation enables clinicians to reduce their workload. Although a detailed efficiency study is deferred to another phase, it is estimated that the Text-Swin-UMamba model takes, on average, 0.518 seconds to produce a segmentation mask for a single 2D slice. This time also includes image loading and text processing. Manual contouring usually takes several tens of seconds per lesion, and even more for irregular lesions. Therefore, a mere partial automation could significantly improve lesion evaluation speed in clinical routine, especially in multi-lesion or longitudinal studies.

## 7. CONCLUSION

The proposed Text-Swin-UMamba model incorporated the Text Tower embedding mechanism into the Swin-UMamba multi-scale decoder to facilitate lesion segmentation. Our model achieved the best segmentation results with a mean Dice score of 82.64% and HD error of 6.34 pixels. It surpassed a similar model with text embedding, LanGuideMedSeg, by 37.79% ($p < 0.001$), while also posting moderate and consistent gains over purely image-based segmentation models, such as xLSTM-UNet (2.58%, $p < 0.001$) and nnUNet (1.01%, $p > 0.05$).

Our results indicate that incorporating short-form text descriptions of findings from the radiology reports can yield a noticeable advantage for lesion segmentation. The clinical context provided by both the imaging and text features guided the network to yield good segmentation results. However, our work has limitations. First, the experimental design utilizes the ULS23 DeepLesion dataset, where lesions are centered on 256×256 images to simplify the segmentation task relative to the original 512×512 images from the original DeepLesion dataset. Such specific design for the dataset constrains the generalizability of our proposed model to more realistic scenarios, where lesions are small compared to surrounding anatomy. Second, other architectural variations, such as a decoder for long-form text generation or relational memory modules[13], have not been investigated due to their implementation complexity. Third, our approach is currently designed only for 2D images. While it can be extended to 3D volumes, this has not been attempted in this pilot work. In summary, Text-Swin-UMamba presents a simple approach to integrate language-guided reasoning for lesion segmentation. The text embedding mechanism can be applied to any nnU-Net model (and its derivatives). We open-source our dataset and code for further research. Future work is targeted towards scaling the approach to full-size CT volumes and enabling long-form descriptions of findings to pre-fill into the reports [22], which may enhance the explainability of lesion segmentation.

## ACKNOWLEDGEMENTS

This work was supported by the Intramural Research Program of the NIH Clinical Center (project number 1Z01 CL040004), National Library of Medicine, and Center for Information Technology. The research used the high-performance computing facilities of the NIH Biowulf cluster.